\documentclass{article}

\usepackage{microtype}      
\usepackage{graphicx}
\usepackage{subfigure}
\usepackage{booktabs}       
\usepackage{algorithmic}

\usepackage{amsmath}
\usepackage{amsfonts}       
\usepackage{nicefrac}       
\usepackage{xcolor}
\usepackage{multirow}

\usepackage{enumitem}

\newsavebox{\fmbox}
\newenvironment{fmpage}[1]
{\begin{lrbox}{\fmbox}\begin{minipage}{#1}}
{\end{minipage}\end{lrbox}\fbox{\usebox{\fmbox}}}

\newenvironment*{packed_itemize}[1][]{
\begin{itemize}[#1]
  \setlength{\itemsep}{4pt}
  \setlength{\parskip}{0pt}
  \setlength{\parsep}{0pt}
}{\end{itemize}}

\newcommand{\floatspace}{\vspace*{-0.05in}}
\usepackage{hyperref}
\usepackage{url}            

\usepackage[accepted]{arxiv2019}

\sysmltitlerunning{No Multiplication? No Floating Point? No Problem!
  Training  Networks for Efficient Inference}

\begin{document}

\twocolumn[
\sysmltitle{No Multiplication? No Floating Point? No Problem! \\
  Training  Networks for Efficient Inference}

\begin{sysmlauthorlist}
\sysmlauthor{Shumeet Baluja}{goog}
\sysmlauthor{David Marwood}{goog}
\sysmlauthor{Michele Covell}{goog}
\sysmlauthor{Nick Johnston}{goog}
\end{sysmlauthorlist}

\sysmlaffiliation{goog}{Google Research, Mountain View, California, USA}

\sysmlcorrespondingauthor{Shumeet Baluja}{shumeet@google.com}

\sysmlkeywords{Machine Learning, SysML}

\vskip 0.3in

\begin{abstract}

  For successful deployment of deep neural networks on highly resource constrained
  devices (hearing aids, earbuds, wearables), we must simplify
  the types of operations and the memory/power resources required during
  inference.  Completely avoiding inference-time floating point operations is one of
  the simplest ways to design networks for these highly constrained environments.
  By quantizing both our in-network non-linearities and our network weights,
  we can move to simple, compact networks without floating point operations,
  without multiplications, and without non-linear function computations.  Our approach
  allows us to explore the spectrum of possible networks, ranging
  from fully continuous versions down to networks with bi-level
  weights and activations.  Our results show that quantization can be done
  with little or no loss of performance
  on both
  regression tasks (auto-encoding) and multi-class classification tasks (ImageNet).
  The memory needed to deploy our quantized networks is less than one-third of the equivalent
  architecture that uses floating-point operations.
  The activations in our networks emit only a
  small number of 
  predefined,
  quantized
  values (typically 32)
  and all of the network's weight
  are drawn from
  a small number of unique values (typically 100-1000) found by
  employing a novel periodic adaptive clustering step during training.

\end{abstract}
]

\printAffiliationsAndNotice{}

\section{Introduction}
\label{sec:intro}

Almost all recent neural-network training algorithms rely on
gradient-based learning.  This has moved the research field
away from using discrete-valued inference, with hard thresholds,
to smooth, continuous-valued activation functions~\cite{werbos1974beyond,rumelhart1986}.
Unfortunately, this causes inference to be done with floating-point operations,
making it difficult to 
deploy
on an increasingly-large set of
low-cost, limited-memory, low-power hardware
in both commercial~\cite{lane2015early} and research
settings~\cite{bourzac2017speck}.

Avoiding all floating point operations
allows the inference network to realize the power-saving gains
available with fixed-point processing~\cite{xilinx-whitepaper}.
To move fully to fixed point,
we need to quantize both the network weights and the
activation functions.  We can also achieve significant
memory savings, by not just quantizing the network weights, but clustering
{\em all of them} across the entire network into a small number of levels.
With this in place, the memory footprint grows about $\frac{1}{3}$ (or less)
as fast as the unclustered, continuous-weight network size.
Additionally, the relative rates at which our memory footprint will grow is easily
controlled using $|W|$, the number of unique weights.
In our experiments, with $|W| = 1000$, we show that we can meet or exceed the classification performance of
an unconstrained network, using the same architecture and (nearly) the same training process.

While most neural networks use continuous non-linearities, many
use non-linearities with poorly defined gradients, without
impacting the training process~\cite{nair2010rectified,glorot2011deep,goodfellow2013maxout}.
When purely quantized outputs are desired, however, such as with binary
units, a number of additional steps are normally
taken~\cite{bengio2013,tang2013learning,raiko2014techniques,courbariaux2016,maddison2016,hou2016loss}
or evolutionary strategies are used~\cite{plagianakos2001training}. At a
high level, many of the methods employ a stochastic binary unit and
inject noise during the forward pass to sample the units and the
associated effect on the network's outputs.  With this estimation, it
is possible to calculate a gradient and pass it through the network.
One interesting benefit of this method is its use in generative
networks in which stochasticity for diverse generation is
desired~\cite{raiko2014techniques}. ~\cite{raiko2014techniques} also
extended~\cite{tang2013learning} to show that learning with stochastic
units may not be necessary within a larger
deterministic network.

A different body of research has focused on quantizing and clustering
network weights~\cite{yi2008new,courbariaux2016,rastegari2016,deng2017,wu2018training}.
Several existing weight-quantization methods
(e.g.,~\cite{courbariaux2016}) liken the process to
Dropout~\cite{srivastava2014dropout} and its regularization
effects. Instead of randomly setting activations to zero when
computing gradients (as with dropout), weight clustering and binarization
tends to push extreme weights partway back towards zero.
Additional related work is given in the next section.

\section {Training Networks for Efficient Inference}

In this section, we separately consider the tasks of (1) quantizing
the output of each unit and (2) reducing the set of allowed weights to
a small, preset, size.  The effects of each method are examined in
isolation and together in Section~\ref{experiments}.

\subsection {Activation Quantization}
\label{tanh}

Figure~\ref{fig:sudounits} gives a simple procedure that we use for
activation quantization.\footnote{
  For the research reported in this paper, we use equal step sizes in
  the activation-output space (see Figure~\ref{fig:sudounits}).
  However nothing in our final system requires this.}
To make this section concrete and easily reproducible,
we show the effects of quantization on tanh's
output.
However, we have employed the exact same method to quantize
ReLU-6~\cite{krizhevsky2010convolutional}, rectified-tanh, and sigmoid
among others.

\newlength\myindent %
\setlength\myindent{1em} %
\newcommand\bindent{%
  \begingroup %
  \setlength{\itemindent}{\myindent} %
  \addtolength{\algorithmicindent}{\myindent} %
}
\newcommand\eindent{\endgroup} %

\begin{figure}[t]
  \centering

  \begin{fmpage}{\linewidth}
    \hspace{-0.025\linewidth}

  \begin{algorithmic}
    \small
    \STATE function gammaD (input, levels):
    \bindent
    \STATE y  $\gets$ $\gamma (input)$
    \STATE $\gamma_{min}$ $\gets$ $\min \gamma(x)$ $\forall x$
    \STATE $\gamma_{max}$ $\gets$ $\max \gamma(x)$ $\forall x$
    \STATE step $\gets$ $ (\gamma_{max} - \gamma_{min}) / (levels - 1)$
    \STATE quant\_y $\gets$ $(\left \lfloor{ (y - \gamma_{min}) / step + 0.5 }\right \rfloor * step + \gamma_{min}$
    \STATE return quant\_y
     \eindent    
  \end{algorithmic}
  \end{fmpage}
  \vspace{0.05in}   

  \includegraphics[width=\linewidth]{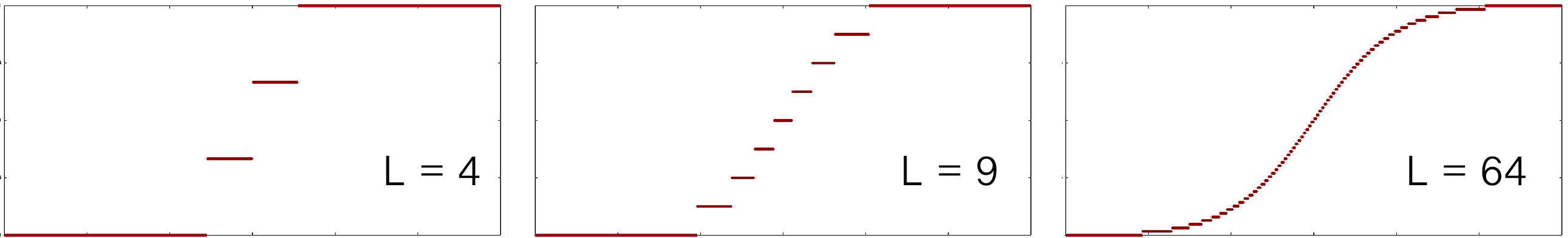}  

\floatspace

\caption{ Quantized non-linearities (detailed for
  reproducibility).  Versions of tanhD (quantized tanh) are shown with
  4, 9, and 64 levels. In the
  largest slope areas of the underlying tanh function, the
  quantization levels change the fastest.  There is no requirement
  to constrain $L$ to a power of 2, though it may be preferred.}
\label{fig:sudounits}

\floatspace

\end{figure}

Naively backpropagating errors with these quantized tanh (tanhD)
units will quickly run into problems as the activations are both
discontinuous and characterized by piece-wise constant functions.  In
order to use gradient-based methods with tanhD units, we need to
define a suitable derivative.  We simply use the derivative of the
underlying function --- (e.g. for $tanhD(x)$, we used $ 1.0 -
tanh^2(x)$).  In the forward pass, both in training and inference, the
output is quantized to $L$ levels.  In the backward pass, we proceed
by ignoring the quantization and instead compute the derivatives of
the underlying function.  Whereas previous studies that attempted
quantization to binary-output units experienced difficulties in
training, we have found that as $L$ is increased, even to relatively
small values ($L \ge 16$), all of the currently popular training
algorithms perform well \emph{with no modification}~\cite{baluja2018},
(e.g. SGD, SGD+momentum, ADAM, RMS-Prop, etc).  A number of studies
have used similar approaches, often in a binary setting (e.g. straight
through estimators~\cite{hinton2012lecture,bengio2013,rippel2017real});
most recently,~\cite{agustsson2017,mentzer2018} used
a smooth mixture of
the quantized and underlying function in the backwards pass.

Why does ignoring the quantization in the backwards pass work?  If
we had tried to use the quantized outputs, the plateaus would not
have given usable derivatives.  By ignoring the quantizations, the
weights of the network \emph{still move in the desired directions}
with each backpropagation step.  However, unlike non-quantized
units, any single move may not affect the unit's output.  In
fact, it is theoretically possible the entire network's output may not
change despite all the weight changes made in a single step.
Nonetheless, in a subsequent weight update, the weights will again be
directed to move, and of those that move in the same direction, some
will cause a unit's output to cross a quantization threshold. This
changes the unit's and, eventually, the network's output.  Further,
notice that for tanhD, Figure~\ref{fig:sudounits}, the plateaus are
\underline{not} equally sized.  Where the magnitude of the derivative
for the underlying tanh function is maximum is where the plateau is
the smallest.  This is beneficial in training since the unit's output
changes most rapidly where the derivative of tanh changes the most
rapidly.

\begin{figure}
  \centering
  \includegraphics[width=\linewidth]{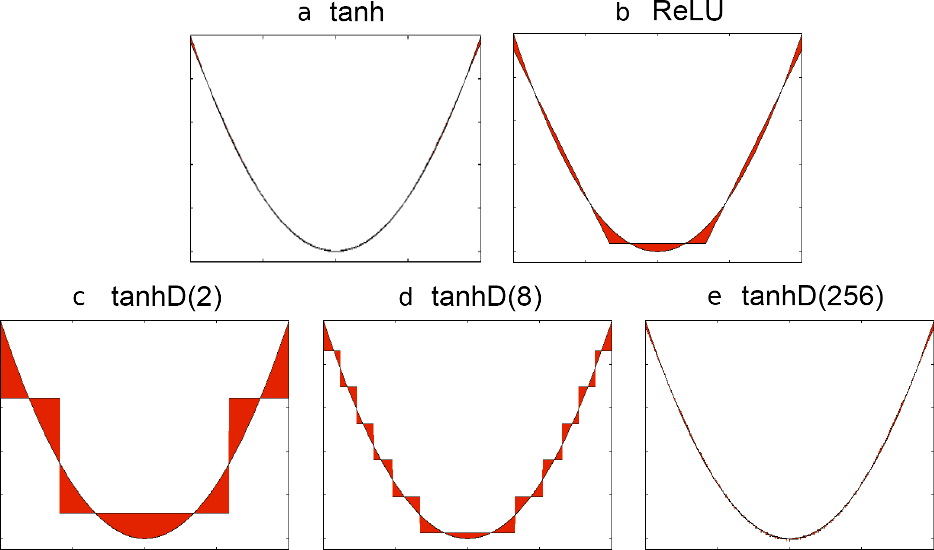}

  \floatspace

  \caption{\footnotesize Fitting a parabola with 2 hidden units.
      The red
  area is the error between the actual and predicted after 100,000
  epochs.  In the top row, the hidden unit activations are standard
  tanh and ReLU.  In the second row, they are quantized versions
  of tanh: tanhD(2), tanhD(8), tanhD(256).  The quantization levels
  clearly affect the
  network's performance.  For example, with tanhD(2), the network has
  found a reasonable symmetric approximation but, with only two hidden
  units it cannot overcome the quantization artifacts. }
    \label{fig:parabola}    

    \floatspace

\end{figure}

Finally, to provide an intuitive example of how these units perform in
practice, see Figure~\ref{fig:parabola}. This shows how a parabola is
fit with a variety of activations and quantization.  In this
example, a tiny network with a single linear output unit and only two
hidden units is used. 
The most revealing graphs are the training curves
with tanhD($L=2$) (Figure~\ref{fig:parabola}-c).  The fit to
the parabola matches closely with intuition; the different levels of
quantization symmetrically reduce the error in a straightforward
manner.  As $L$ is increased (d and~e), the performance
approximates and then matches the networks trained with tanh and ReLU
activations.

To summarize, a simple procedure to quantize the outputs of a unit's
activations
was given.  For ease of
exposition and clarity, it was presented with tanh, though testing has
been done with most, if not all, the commonly used activation
functions.  Beyond tanhD, we will demonstrate the use of quantized
ReLU activations in Section~\ref{experiments}.

\subsection {Weight Quantization}
\label{weight clustering}

We turn now to reducing the set of allowed weights to a small,
predefined, number.  As the goal of our work is efficient inference;
we do not attempt to stay quantized \emph{during} the training
process.  The process used to obtain only a small number of unique
weights is a conceptually simple addition to standard network
training.  Like activation quantization, it can be used with any
weight setting procedure - from SGD or ADAM to evolutionary algorithms.

Previous
research has been conducted in making the network
weights integers~\cite{yi2008new,wu2018training}, as well as reducing
the number of weights to only binary or ternary
values~\cite{courbariaux2016,deng2017} during both testing and
training, using a stochastic update process driven by the sign bit of
the gradient.
Empirically, many of the previous techniques that either compress or
quantize weights on an already trained model perform poorly on
real-valued regression problems. While our implementations
of~\cite{denton2014,nvidia2016,qualcomm2017} are quite successful on
classification problems, we were unable to achieve comparable
performance using those techniques on networks that perform image
compression and reconstruction, using the architectures described
in~\cite{balle2016}.  We hypothesize that the reason for this is that
quantizations can create
sharp cuts that seem to be beneficial for decision boundaries but
hinder performance when regressing to real-valued variables.  Tasks in
which real-valued outputs are required have become common recently
(\emph{e.g.} image-to-image-translation~\cite{isola2016image}, image
compression and speech
synthesis, to name a few).
Fortunately, our method exhibits good performance on regression tasks,
as well as providing an easily tunable hyper-parameter
(the number of
weight clusters), thereby alleviating any remaining
task impact.

Perhaps the most straightforward approach to creating a network with
only a limited number of unique weights ($|W|$) is to start with a
trained network and
quantize the
weights
to a small number, $|W|$, of
equally spaced
values.
While this clearly allows us to use fixed-point arithmetic,
this approach has
limitations.  The number of uniformly spaced levels must be large to
retain good performance: \cite{qualcomm2017} showed that, even with
1024 distinct (equally-spaced) levels on the first layer of Alexnet,
straightforward quantization (without post-quantization fine tuning)
resulted in a classification accuracy drop of 72\%.\footnote{This
  number is based on~\cite{qualcomm2017} Figure~4 ``optimized
  bit-width'' with conv1 weights
  having $2^{10}$ distinct values; conv2 and conv4 each having $2^5$
  distinct values; and conv3, conv5, fc1 and fc2 each having
  $2^6$ distinct values.  The total number of
  distinct weight values within the network would be 1344.}
As will be explained in
Section~\ref{Implementation Details}, moving
away from uniform
quantization does \emph{not} force us away from fixed-point
arithmetic: we are able to reap the power/computational benefits of
this regime without resorting to uniform sampling.

\begin{figure}
\centering

  \includegraphics[width=\linewidth]{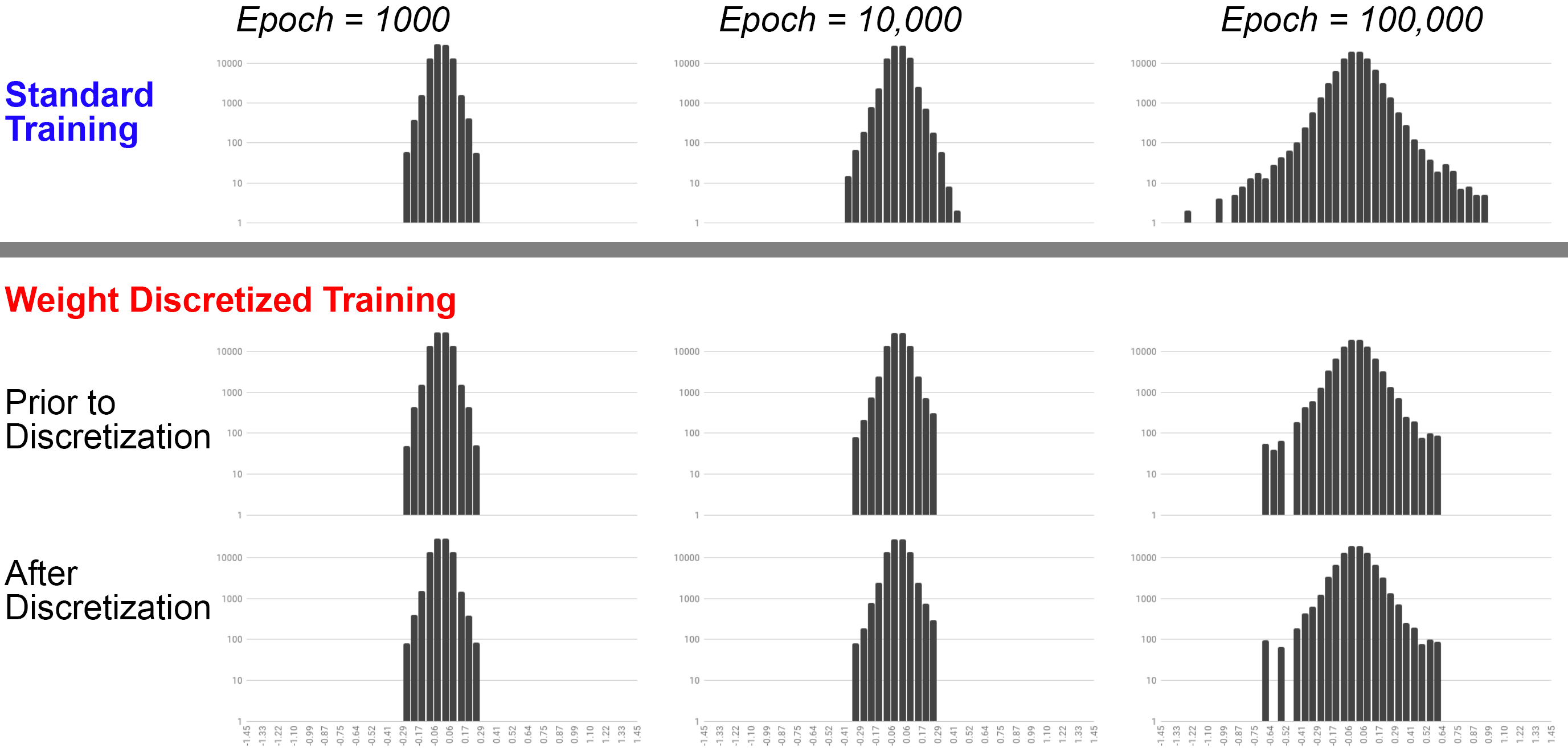}

\caption{ (top) MNIST distributions of weights trained with
  no weight quantization, shown for epochs 1000,10000 and 100000.  (middle)
  Distribution of weights when trained with weight maximum of 1000
  unique weights.  Same Epochs shown immediately prior to weight
  quantization.  (bottom) Weights shown after weight quantization. All
  y-axes
  are log-scale to show lesser occupied bins.  The frequency
  distributions are nearly Laplacian: they look Gaussian here due to
  the log-scale.}
\label{fig:histograms}      

\floatspace

\end{figure}

To address the limitations in uniform quantization, we instead
\emph{adaptively cluster} the weights
throughout the training process.  Rather than fully training a
network and then quantizing the weights, a recurring clustering step
is added into the training procedure.  Periodically (after every 1000
steps, in our experiments), all of the
weights in the network, including the bias weights, are clustered in a
one-dimensional (the value of the weight/bias) k-means process~\cite{jain2010data}.\footnote{
  All of the clustering approaches that we tried
  (e.g., LVQ~\cite{kohonen1995learning},
  HAC~\cite{duda1995patternHAC}, k-means)
  gave similar
  results.
  We used k-means for simplicity.
}  Clustering, rather than using uniformly sized
bins, ensures that bin spacing respects the underlying distribution of
the weights (Figure~\ref{fig:histograms}).

\begin{figure}
\centering

\includegraphics[width=0.3\linewidth]{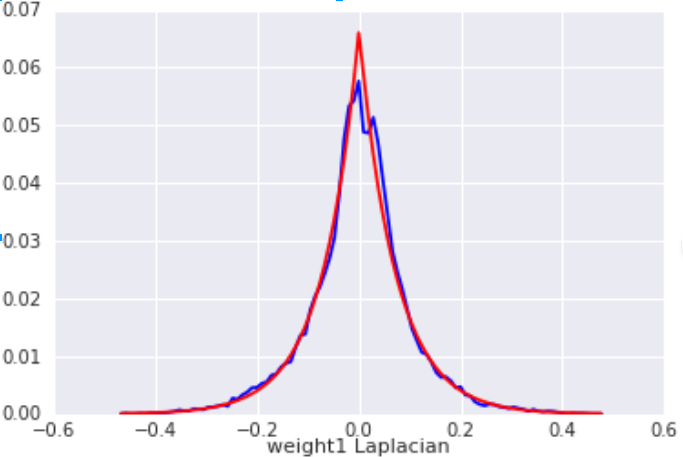}\hfill\includegraphics[width=0.3\linewidth]{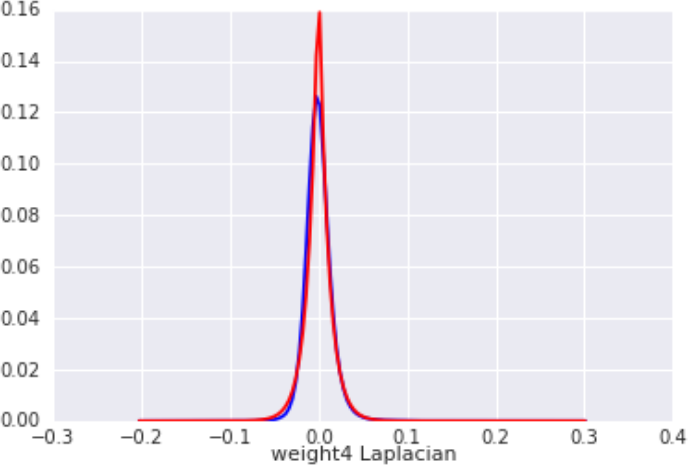}\hfill\includegraphics[width=0.3\linewidth]{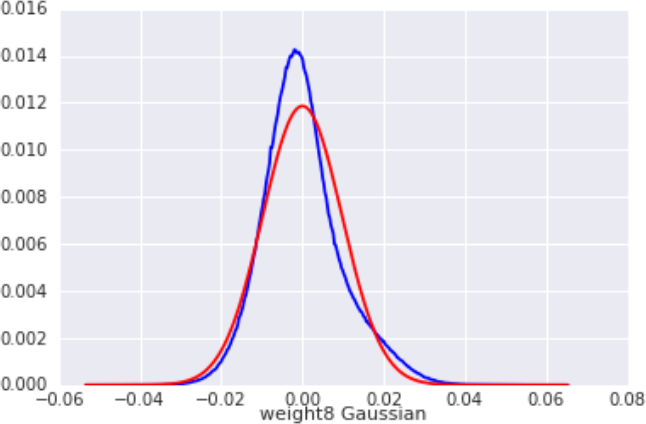}

\caption{Histograms of weights from a well-trained (continuous)
  Alexnet, for the layers 1, 4, and 8.  The blue line is the actual weight
  distribution; the red one is the best fitting distribution (either
  Laplacian or Gaussian).  The distributions of layers 2, 3, and 5
  look very similar to the Laplacian distribution of layer 4.  The
  distributions of layers 6   and 7 are similar to the Gaussian
  distribution of layer 8, but with smaller variances.}
\label{fig:Alexnet histograms}

\floatspace

\end{figure}

As will be seen in our results (Section~\ref{experiments}), using
k-means clustering on the weights and biases of small- and medium-sized
networks works extremely well.  However, k-means clustering on more
than about a million parameters can become prohibitively slow, even though
we only do this clustering once every one thousand steps.  Most modern networks
far surpass a million weights: Alexnet, for example, has more
than 50 million weights and biases.  In Section~\ref{Alexnet},
we use a simple method to sidestep this problem by subsampling in
the weight/bias space, and determine cluster centers using only 2\% of
the parameters.  This subsampling allows for faster training, 
but does not make use of the available information about the
weight/bias distributions
and, as we will see in Section~\ref{Alexnet},
gives about 3\% lower network accuracy.

\begin{figure}
\centering

  \includegraphics[width=0.48\linewidth]{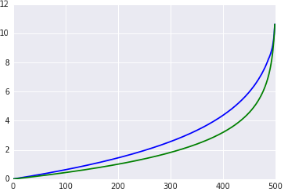}\hfill  \includegraphics[width=0.48\linewidth]{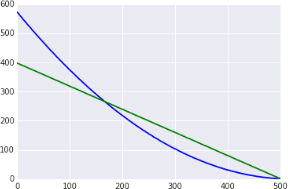}

\caption{Quantization centers (left) and bin counts (right), for the positive range
  of a Laplacian distribution with a standard deviation of $\sqrt{2}$,
  for $|W|=1000$ on 100,000 samples.  The green curve shows the
  centers/counts when minimizing the $L_1$ quantization error; the
  blue curve when minimizing the $L_2$ error.  When minimizing overall
  network quantization
  error, the quantization centers are non-uniformly spaced, with wider
  spacing at large amplitudes, due to the Laplacian distribution
  of weight amplitudes.}
\label{fig:DistributionModel}

\floatspace

\end{figure}

Instead of simply subsampling, we can make use of what we know about fully-trained weight
distributions:  that many resemble Laplacian or Gaussian
distributions.
Figure~\ref{fig:histograms} shows this for a fully-trained
MNIST-classification
network and Figure~\ref{fig:Alexnet histograms} shows it for the
fully-trained Alexnet.  This insight opens interesting model-based
approaches to 
quantization.  Specifically, if the ``natural'' (unquantized)
weight/bias distributions on large networks really should be Laplacian (or
Gaussian), the loss in accuracy of the quantized network might be
traced back to the overall $L_1$ or $L_2$ error of the quantized
weights/biases compared to their ``natural'' distributions.  Under this assumption,
we should be able to determine the optimal quantization levels
based on these parameterized models.  Figure~\ref{fig:DistributionModel} shows
cluster centers and occupancies for Laplacian
distributions that minimize its $L_1$ or $L_2$ error in the
weight/bias space.  While 
this does not necessarily correspond directly to minimizing the
accuracy loss, using the Laplacian $L_1$ not only matched the classification accuracy than we got from unconstrained k-means, but actually surpassed it, as we will show in Section~\ref{Alexnet}.

One interesting characteristic of the $L_1$ Laplacian-based clustering
model is that we can describe the best cluster center locations in
closed form, as a function of the extreme values that were seen in the
sampled set.  Specifically, for minimum $L_1$ error and using an odd
number of cluster centers, the $N$ cluster
centers should be at $a \pm b L_i$ where
$a$ is the mean value of the network weights and $b$ is a scaling
factor and where $L_i = L_{i-1} + \Delta_i$ with $\Delta_i = - \ln(1 -
2 \exp(L_{i-1}) / N)$ and $L_0 = 0$.  We set the scaling factor, $b$,
using the cluster occupancy curve for guidance.  From
Figure~\ref{fig:DistributionModel}, for minimum $L_1$ error on a
fair sample set from a Laplacian distribution, occupancy of the
clusters should fall linearly.  However, early in training, the
observed weight distribution is far from a fair Laplacian sample: many more
weights are near zero than dictated by a Laplacian model.  We adjust
for this by setting $b$ so that, early in training, the maximum
quantization level $a \pm b L_{N/2}$ is at or slightly beyond the
maximum observed weight.  Specifically, we start with
$b = W_{max} / L_{N/2}$ where $W_{max}$ is the maximum amplitude
difference between any weight and the mean $a$.  This scaling
allows us to accurately model the largest magnitude weights.

On a pragmatic note, we found that in the beginning of training, the weights
were too tightly clustered around the mean (compare the left and right
columns in Figure~\ref{fig:histograms}).
We determined that by slightly ``nudging'' $b$ in early training,
we could speed-up convergence without losing final network accuracy.  If
$W_{max} < 0.5$, we change $b$ to move $L_{N/2}$ outward by
$\frac{1}{2 (1 - W_{max})} b \Delta_{N/2}$.  Later in training, tying
our scaling parameter to $W_{max}$ means that we would lose the
regularization benefits seen in Figure~\ref{fig:histograms}-b
and~-c. To avoid losing that, we ``nudge'' the value 
of $b$ just slightly lower, by $\frac{1}{4} b \Delta_{N/2}$, whenever
the activation weights are spread out by more than the expected range
of desired values (specifically, whenever $W_{max} > 1.25$).  We will
revisit this Laplacian-model--based approach with our investigation of
Alexnet (Section~\ref{Alexnet}).

\begin{figure*}[t]
  \centering \rotatebox{90}{\footnotesize ~~~~~2 Hidden Layers }  
  \begin{minipage}[t]{0.32\linewidth}
    \includegraphics[width=\linewidth]{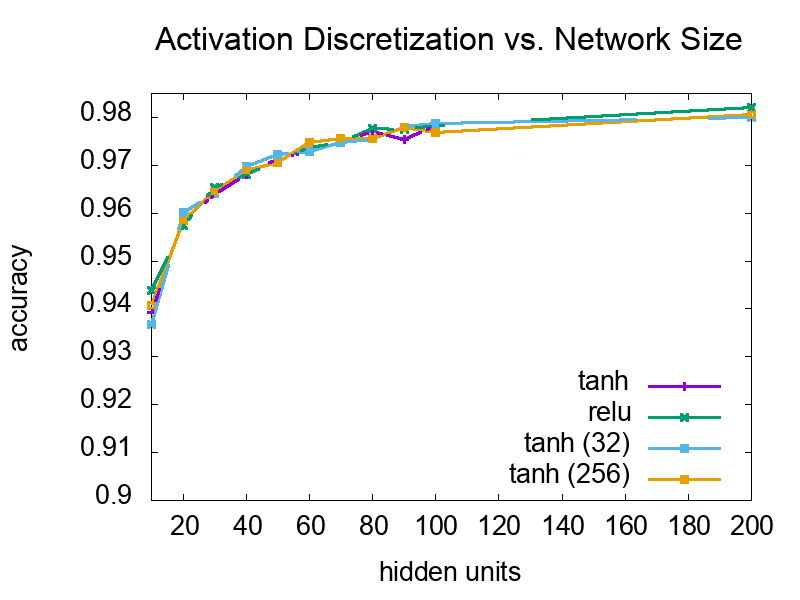}
  \end{minipage}\begin{minipage}[t]{0.32\linewidth}
    \includegraphics[width=\linewidth]{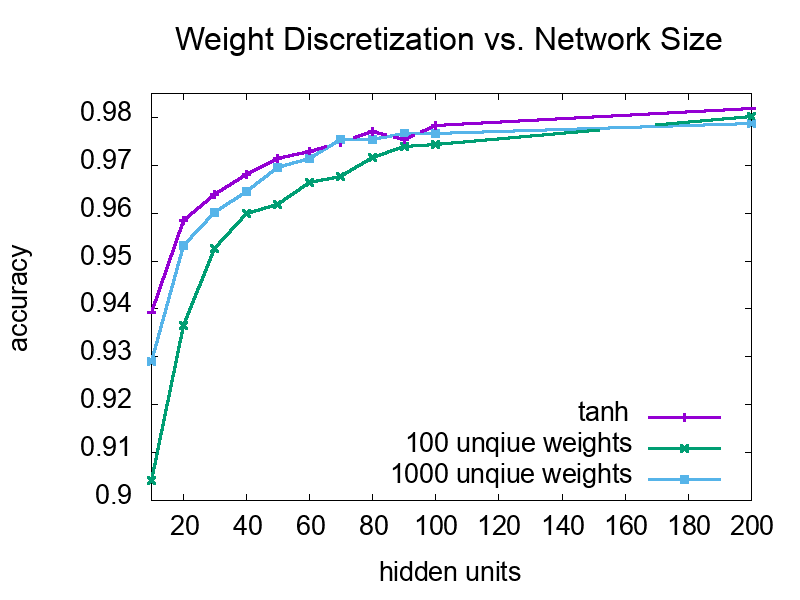}
  \end{minipage}\begin{minipage}[t]{0.32\linewidth}
    \includegraphics[width=\linewidth]{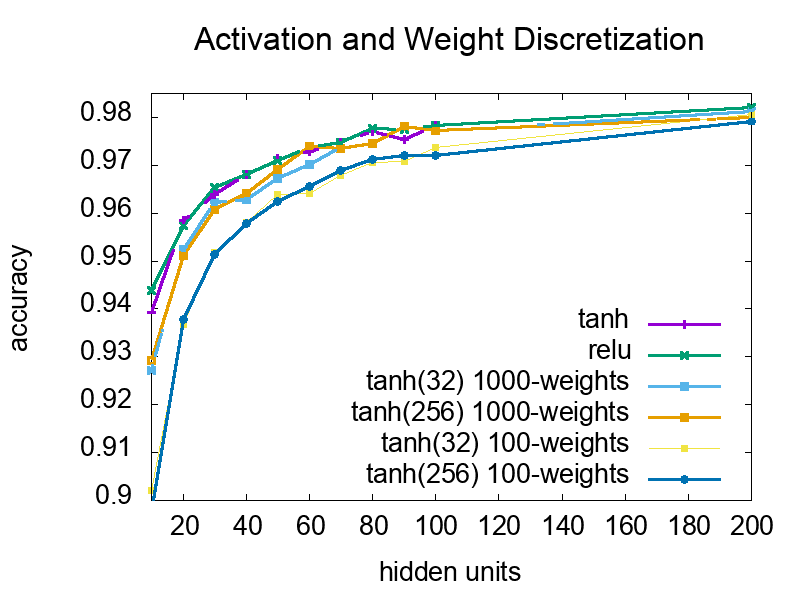}
  \end{minipage}
  \\
  \centering \rotatebox{90}{\footnotesize ~~~~~4 Hidden Layers }    
  \begin{minipage}[t]{0.32\linewidth}
    \centering \footnotesize
    \includegraphics[width=\linewidth]{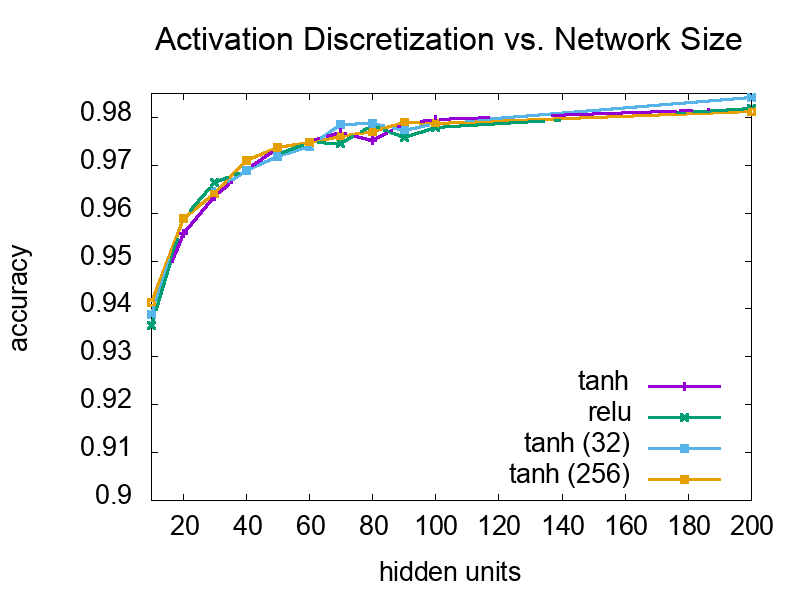} \\ (a) Activations quantized \\
  \end{minipage}\begin{minipage}[t]{0.32\linewidth}
    \centering \footnotesize
    \includegraphics[width=\linewidth]{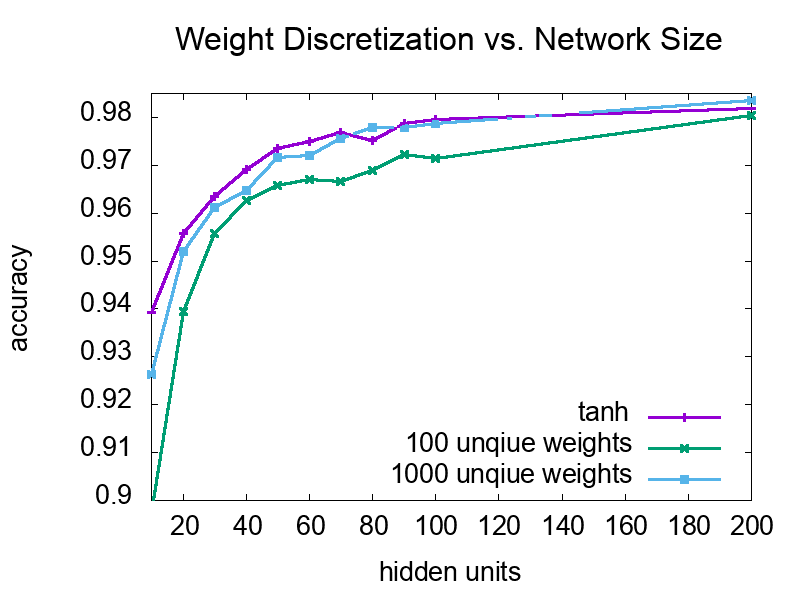} \\ (b) Weights quantized \\
  \end{minipage}\begin{minipage}[t]{0.32\linewidth}
    \centering \footnotesize
    \includegraphics[width=\linewidth]{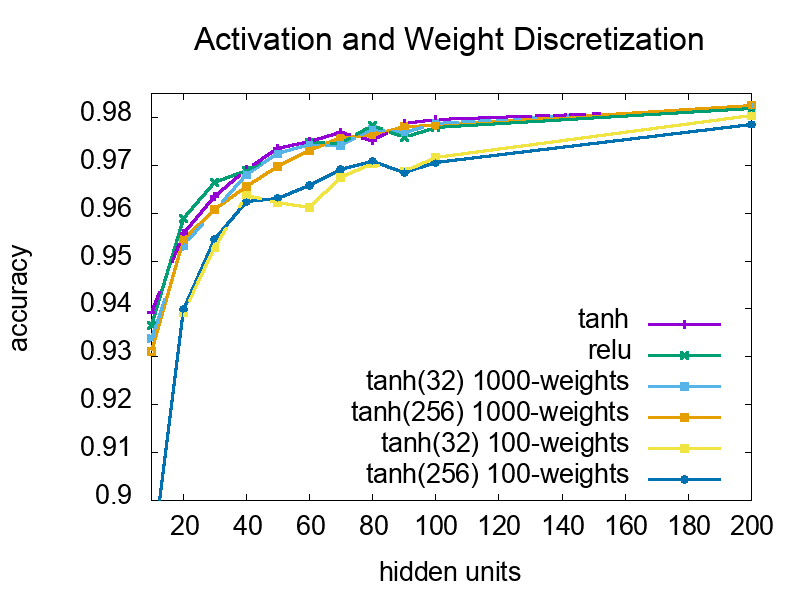}  \\ (c) Both quantized \\
  \end{minipage}
\caption{\footnotesize Effects of activation quantization vs. number of hidden
  units on MNIST classification.  The only performance loss is
  observed when the number of unique weights is reduced to 100. With
  1000 weights, with or without activation-quantization, performance
  matches or surpasses ReLU and tanh.  Average of 3 runs shown.}
\label{fig:mnist}

\floatspace

\end{figure*} 

Whichever way we pick the quantization clusters (whether using k-means or
model-based approaches), once the clusters are
created, each weight is replaced 
with the centroid of its assigned cluster, thereby reducing the number
of unique weights to $|W|$.  After weight replacement, training
continues
with no modifications until the next
clustering step.  At this point, the cycle repeats. For all of our
experiments, quantization occurs after every 1000 training steps.

This procedure, though simple, has some subtle effects.  First, as a
training regularizer, it keeps the range of the weights from growing
too quickly, as there is a persistent ``regression to the mean.''
Second, it provides a mechanism to inject directed noise into the
training process.  As we will show, both of these properties have, at
times, yielded \emph{improved} results over allowing arbitrary valued
weights. Figure~\ref{fig:histograms} (row b) shows the distribution of
weights at the beginning, middle, and end of training when weight
clustering is used, immediately prior to the quantization step.  With
1000 clusters used throughout training, the weights after replacement
(Figure~\ref{fig:histograms}, row c) appear very similar to the
unclustered weights (Figure~\ref{fig:histograms}, row b).

\section {Experiments}
\label{experiments}

We experimented with many tasks and network architectures to
determine
how quantizations of
activations and weights affected
performance.
These experiments
included testing memorization capacity, real-value function
approximation, and numerous classification problems.  Because of space
limitations, however, we only present the three most often researched
tasks; these are representative of the results seen across our
studies.  We present two classification tasks:
MNIST~\cite{lecun1998mnist}, and
ImageNet~\cite{deng2014imagenet}.
We also present
a real-valued output task: auto-encoding images, the crucial building
block to neural-network based image compression.

In all of our tests, we retrained the baselines to eliminate the
possibility of any task-specific heuristics.  In some
cases, this led to lower baseline
performance than state-of-the-art; however,
since our goal is to measure the relative effect of quantizations on
any network, the results provide the insights needed.

\subsection {MNIST}
\label{MNIST}

For MNIST, we train a fully connected network with
ADAM~\cite{kingma2014adam} and vary the number of hidden units to
explore the trade-off between quantization, accuracy, and network
size.
These networks ranged from just over 8,000 parameters up to nearly
110,000 parameters, depending on the architecture and number of
hidden units.
Figure~\ref{fig:mnist}-a contains the
performance of the networks using ReLU and tanh activation functions
with no quantizations; these are the baselines.  Since tanh slightly
outperformed ReLU, we will quantize tanh in our experiments.

\begin{figure*}
  \centering \rotatebox{90}{\scriptsize ~~~~~Convolution }~
  \begin{minipage}[t]{0.32\linewidth}
  \includegraphics[width=\linewidth]{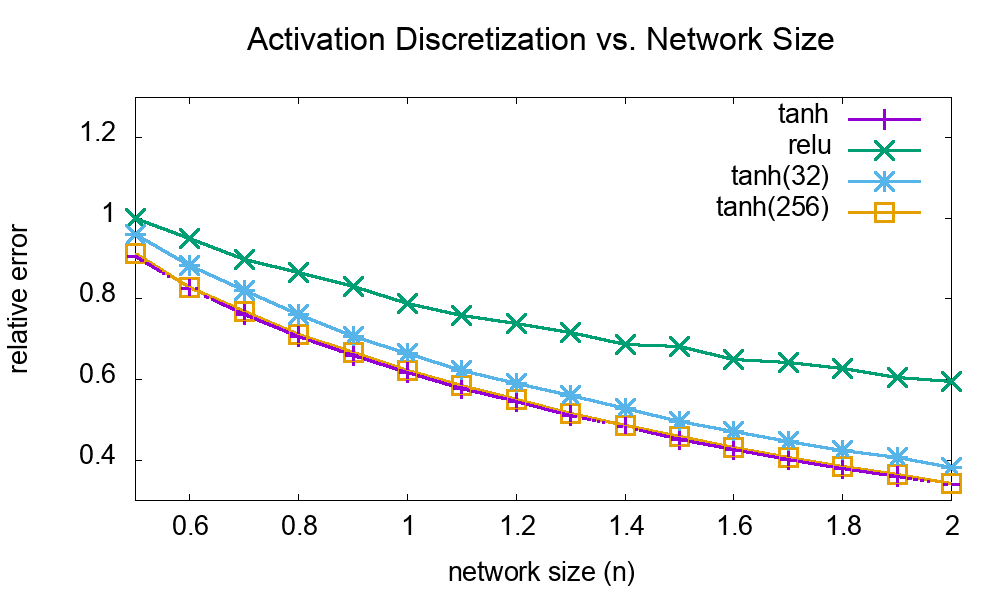}
  \end{minipage}\begin{minipage}[t]{0.32\linewidth}
  \includegraphics[width=\linewidth]{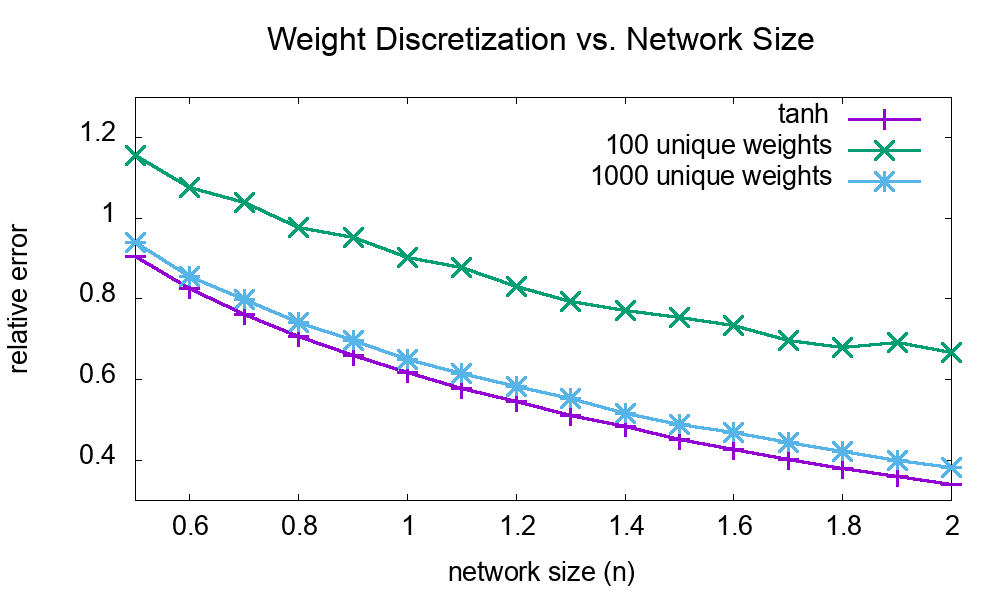}
  \end{minipage}\begin{minipage}[t]{0.32\linewidth}
  \includegraphics[width=\linewidth]{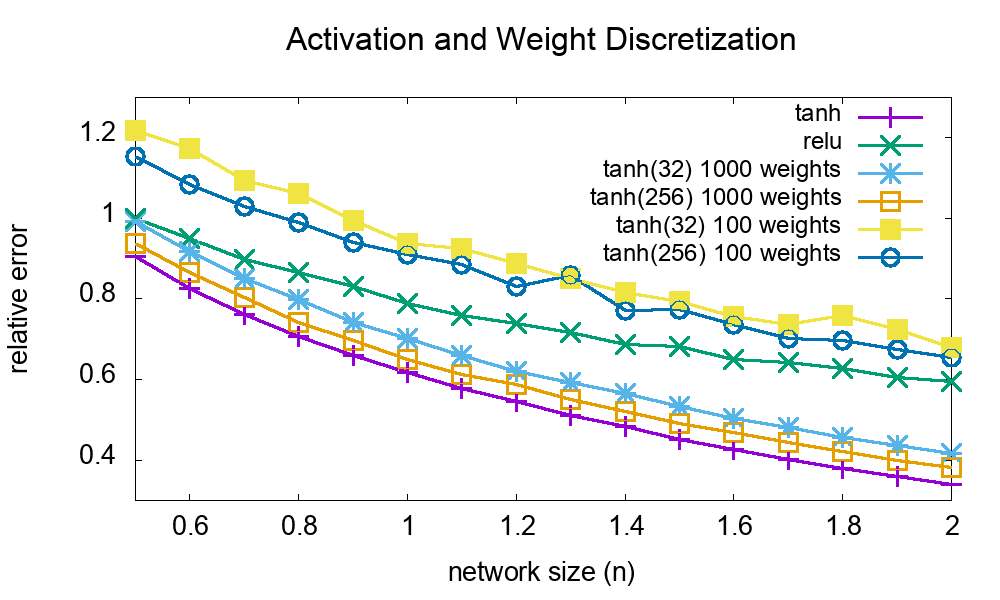}    
  \end{minipage}
  \\
  \centering \rotatebox{90}{\scriptsize ~~~~~Fully~Connected }~
  \begin{minipage}[t]{0.32\linewidth}
    \centering \footnotesize
    \includegraphics[width=\linewidth]{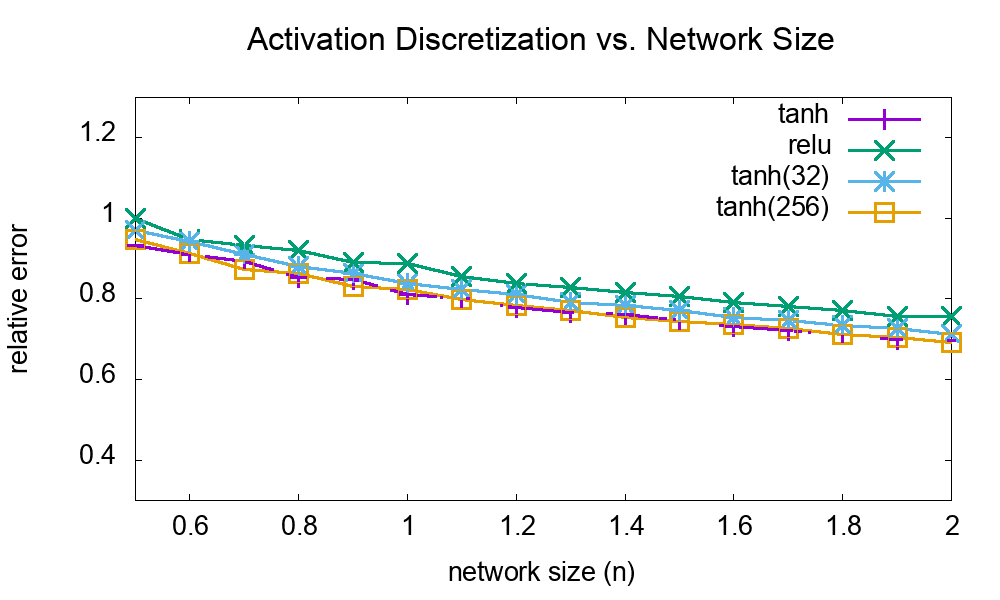} \\ (a) Activations quantized \\
  \end{minipage}\begin{minipage}[t]{0.32\linewidth}
    \centering \footnotesize
    \includegraphics[width=\linewidth]{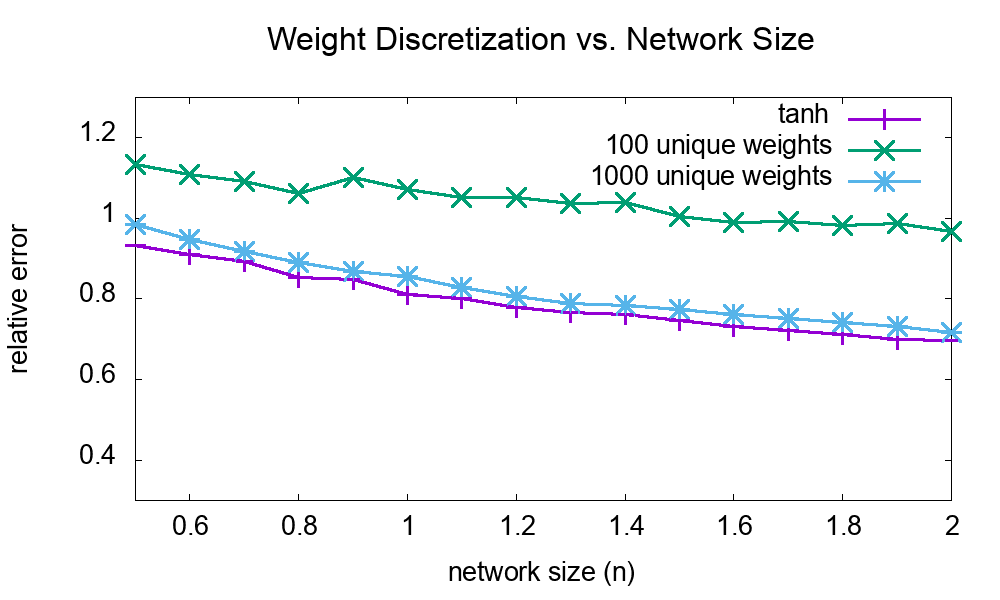} \\ (b) Weights quantized \\
  \end{minipage}\begin{minipage}[t]{0.32\linewidth}
    \centering \footnotesize
    \includegraphics[width=\linewidth]{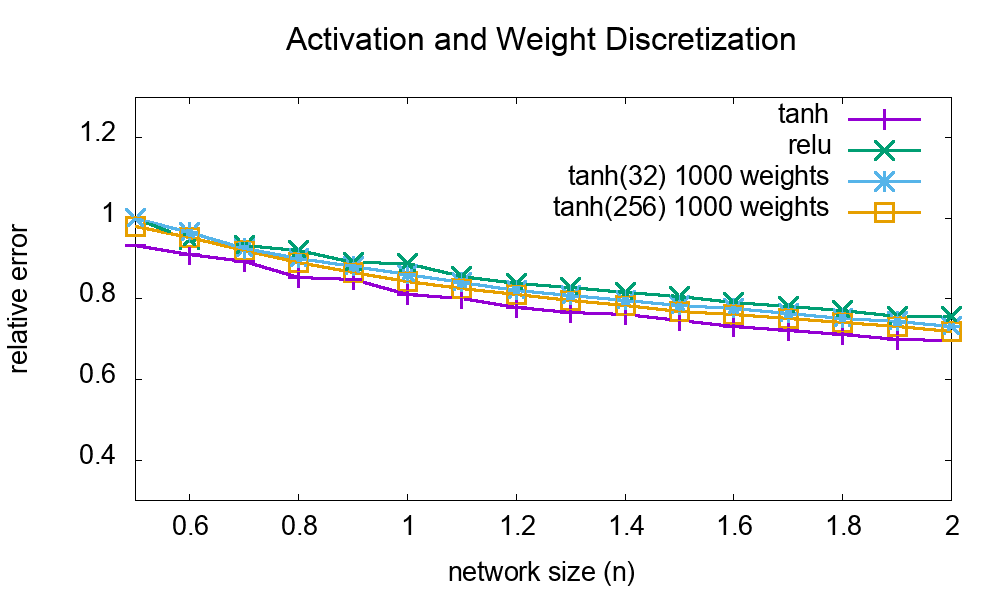} \\ (c) Both quantized \\
  \end{minipage}

  \caption{\footnotesize (a) Effects of activation quantization
    vs. number of hidden units on auto encoding. The worst performing
    is unmodified ReLU. Tanh and tanhD(256) performed better and were
    equivalent to each other. (b) When weights are quantized,
    performance declines as $|W|$ decreases. (c) Combined effects of
    both quantizations ($|W|=100$ is not shown at bottom of (c), due
    to plot range).}

\label{fig:autoencode}

\floatspace

\end{figure*} 

First, we examine the effect of only quantizing each unit's
activations.  We experimented with 8 sets of activation quantization
(2, 4, 8, ... 256 levels). We found that both tanhD(8) and tanhD(16)
often perform as well as tanh and ReLU in performance when there are
$>3$ hidden units per layer. At tanh(32) and above,
performance is largely indistinguishable from tanh
(Figure~\ref{fig:mnist}-a).  Next, we examine weight
quantization in isolation using k-means clustering.  Two sets of experiments are presented:
$|W|=100$ and $|W|=1000$ (Figure~\ref{fig:mnist} b). With 1000 unique weights allowed, the
performance is nearly identical to no weight quantization.  However,
when $|W|$ is reduced to 100, there is a decline in performance.
Nonetheless, note that even with $|W|=100$ the performance recovers
with additional hidden units -- hinting towards the trade-off in
representational capacity between number of distinct values a weight
can represent and the number of weights in the network.

Finally, when both quantizations are combined, we again see that the
only noticeable negative effect comes when the number of unique
weights is set to 100.  No matter which activation is used, when 100
weights are used, performance decreases.  This same trend holds true
for networks with a depth of 2 hidden layers (top row) and with 4
hidden layers (bottom row).

\subsection {Auto-Encoding}
\label{autoencoding}

A number of recent as well as classic research papers have used
auto-encoding networks as the basis for image
compression~\cite{cottrell1988principal,kramer1991nonlinear,jiang1999image,toderici2015,Svodboda2016,toderici2017full}.
To recreate the input image, real values are used as the outputs.  As
discussed earlier, real value approximation can be a more challenging
problem domain than classification when quantizing the network.

For these experiments, we train two network architectures:
convolutional
(26,000 to 380,000 parameters)
and
fully connected
(160,000 to 660,000 parameters).  ADAM is
used for training, and $L_2$ error is minimized. We trained with the
ImageNet train-set and all tests are performed with the validation
images.  The smallest conv-network has four $2\times2$ conv.
layers with ($50 n$, $50 n$, $40 n$, $20 n$) depth, followed by 3
conv-transpose layers with depth ($40 n$, $50 n$, $50 n$).  The last
two layers are $1\times1$ conv. with depth 20 and 3.  For the
second experiment, the fully-connected network has 7 hidden layers
with ($50 n$, $50 n$, $40 n$, $20 n$, $40 n$, $50 n$) units each.  To
examine the effects of network size, $n$ is varied from 0.5 to 2.0 for
both experiments.

Because the raw numbers are not meaningful in isolation, we show
performance measurements relative to training the smallest network
with ReLU activations and no quantizations (the graphs for both
architectures can be compared to see effects of network size).  In
Figure~\ref{fig:autoencode}-a for both architectures, ReLU performed
worse than tanh.  TanhD(32) and TanhD(256) tracked the performance of
tanh closely for all sizes of the network.  Similarly to the MNIST
experiment (Figure~\ref{fig:mnist} b), reducing the number of weights
to $|W|=100$ hurt performance.  With $|W|=1000$, the performance
decline was much smaller; however, unlike with MNIST, there was a
discernible effect.

\begin{table*}
  \caption{AlexNet Experiments. Results are sampled from the step
   with the highest recall@5, up to step 1,000,000 (i.e., early
      stopping is allowed).  Results marked with $^*$ are based on
      training twice and picking the training run with the best
    recall@5.  Recall@1 and @5 are always sampled from
    the same training run and the same step.
    For the results in the \emph{quantized inputs} columns,
    the network-input pixel values were
    quantized to the same number of levels used for activation 
    quantization.}
  \centering
  \footnotesize
  \begin{tabular}{p{2.7in} c||cc|cc|cc}\hline
    \multicolumn{2}{c||}{ } &  Activation & Unique &
    \multicolumn{2}{c|}{Unquantized inputs} &
    \multicolumn{2}{c}{Quantized inputs} \\\cline{5-8}
    
    Experiment & \# & Levels  & Weights & Recall & Recall & Recall & Recall \\
    
    \multicolumn{2}{c||}{ } &  ($|A|$) & ($|W|$) & @1  &  @5 & @1  &  @5 \\\hline\hline

    AlexNet w/ ReLU  & 0
    & -    & - & 57.4 &  80.4 & - & - \\
    AlexNet w/ ReLU6  & 1
    & -    & - & 56.4 &  79.8 & - & - \\\hline

    \multirow{4}{*}{\parbox{2.4in}{Continuous weights, only quantized
        activations.}}
    & 2 & 256  & - & 56.1 & 79.8 & - & - \\
    & 3 & 32 & - & 56.0 & 79.4 & 56.0 & 79.6 \\
    & 4 & 16 & - & 55.8 & 79.4 & 55.4 & 78.9 \\
    & 5 & 8 & - & 53.4 & 77.7 & 52.6 & 77.1 \\\hline

    \multirow{2}{*}{\parbox{2.3in}{k-means quantized weights and quantized
      activations (no dropout).}}
    & 6  & 32 & 1000 & 52.5$^{*}$ & 76.3$^{*}$ & 52.1 & 76.0 \\
    & 7 & 32 & 100 & 48.6$^{*}$ & 73.1$^{*}$ & 47.2 & 72.2 \\\hline

     Laplacian quantized weights, quantized
       activations: & & & & & \\
    ~~~~{\em - with dropout}  & 8 & 32   & 1000 & 55.5$^{*}$ & 79.3$^{*}$ & 55.5 &
       79.2 \\
     ~~~~{\em - without dropout} & 9 & 32   & 1000 & 57.1$^{*}$ &
       79.8$^{*}$ & 56.9 & 79.4 \\\hline
    \end{tabular}
  \label{alexTable}

\floatspace

\end{table*}      

When the two quantizations were combined,
again, the largest impact was a result of setting the weight
quantization levels too low.  As before, increasing the network size
returns the performance lost due to weight and activation
quantization.  Importantly, this task indicates that although the
quantization procedures do indeed take a larger toll on the
performance with real-outputs, quantization remains a viable
approach for network computation/memory reduction.  The amount of
performance degradation tolerated can be explicitly dictated by the
needs of the application by controlling $|W|$.
It is worth repeating here that while our implementations of many of the recent competing methods in~\cite{denton2014,nvidia2016,qualcomm2017} are quite successful on
classification problems, we were unable to achieve comparable
performance using those techniques on compression and reconstruction.

\subsection{AlexNet}
\label{Alexnet}

To evaluate the effects of quantization on a larger network (more
than 50 million parameters), we used
AlexNet~\cite{krizhevsky2012imagenet} to address the 1000-class
ImageNet task.  To ensure that we are training our network correctly,
we first retrained AlexNet from scratch using the same architecture
and training procedures specified in in~\cite{krizhevsky2012imagenet};
some small differences are: we employed an RMSProp optimizer, weight
initializer sd=0.005, bias initializer sd=0.1, one Nvidia Tesla P100
GPU, and a stepwise decaying learning rate.  Our network achieved a
recall@5 accuracy of 80.4\% and recall@1 accuracy of 57.4\%.  This
should be compared to the accuracy reported in
~\cite{krizhevsky2012imagenet} of 81.8\% and 59.3\%, recall@5 and
recall@1, respectively.  The small difference in performance was
because we did not use the PCA pre-processing,
which~\cite{krizhevsky2012imagenet} cite as causing approximately the
difference seen.
All of the remaining comparisons
will use the exact same training procedure, only differing in which
quantizations and activations are used.

To begin, we examined the effect of switching to ReLU6 instead of
ReLU.  AlexNet with ReLU6 achieved a recall top-1 of 56.4\% and top-5
of 79.8\% (Experiment \#1).  This change is needed to support
activation quantization, to give a bounded range.
With that change, we then
examine the performance of only quantizing
the activations and inputs, without weight quantization, 
in Experiments \#2-\#5 \footnote{
  We separately report two approaches: first, quantizing activation outputs
  without quantizing the network inputs  and, second, quantizing both
  activation outputs and network inputs to the same number of levels.  This
  allows us to compare to other prior work which left the first and
  last layers of their networks unquantized~\cite{Zhou2016,rastegari2016}.}.
using 256 activation levels (8 bits) down to
only 8 levels (3 bits).  In Experiments \#2 and~\#3, there is little
degradation in performance in comparison to using the full 
floating point (32 bits) (Experiment \#1).  Below 32 levels with both input and
activation quantization (right most columns in
Table~\ref{alexTable}), performance declines (Experiment~\#4 and~\#5).

Using the most aggressive acceptable activation quantization (32 values), we
turn to our next experiment: reducing the number of unique weights
allowed.  We set $|W|=1000$ (Experiment \#6). The only training
modification was the elimination of dropout.  As illustrated in
Figure~\ref{fig:histograms}, the quantization
process itself works as a regularizer, so  dropout is not needed.
\cite{wu2018training} took a similar approach and removed
dropout from their AlexNet quantization
experiments. It should also be noted that \cite{wu2018training}
  \emph{did not} quantize the last layer's weights for reporting results.
  All of our quantized AlexNet results include quantization
  of the final layer's weights. Also, \cite{wu2018training} did not 
  quantize the network inputs
  -- we show performance with input quantization (rightmost columns ``Quantized Inputs''). 
For speed in training, only a
randomly selected 2\% sample of the full weight/bias set was used for k-means
clustering; however, all the weights/bias
parameters were then set to those cluster centers.

\begin{table*}
\begin{center}
\caption{Accuracy of Alexnet under Quantization: Comparison to
  Prior Work}
\label{table: compare}
\begin{tabular}{|l|c|c|c||c|c|c|}\hline
  & \multicolumn{3}{c||}{Recall@1} & \multicolumn{3}{c|}{Recall@5} \\\hline
 & baseline & quantized & difference & baseline & quantized & difference \\\hline

Our work & {\bf 57.4\%} & {\bf 57.1\%} & {\bf -0.3\%} & 80.4\% &
    {\bf 79.8\%} & {\bf -0.6\%} \\\hline
WAGE~\cite{wu2018training} & - & - & - & {\bf 80.7\%} & 75.9\% & -4.8\% \\\hline
DoReFa~\cite{Zhou2016} & 55.9\% & 53.0\% & -2.9\% & - & - & - \\\hline
QNN~\cite{Hubara2016} & 56.6\% & 51.0\% & -5.6\% & 80.2\% & 73.7\% & -6.5\% \\\hline
XNOR-Nets~\cite{rastegari2016} & 56.6\% & 44.2\% & -12.4\% & 80.2\% &
69.2\% & -11.0\% \\\hline
Optimized fixed-point~\cite{qualcomm2017}\footnotemark[2]
  & - & - & - & 80.3\% &
22.6\% & -57.7\% \\\hline
\end{tabular}
\end{center}

\floatspace

\end{table*}

Examining Experiments~\#6 and~\#7,
we see that Experiment \#7,
with only 100 unique weights, performed much better than we would have
expected given its earlier performance in Subsections~\ref{MNIST}
and~\ref{autoencoding}.  We
speculate that unlike the other tasks in which setting $|W|=100$ was
detrimental to performance, AlexNet has so much extra capacity and
depth that the effective decrease in representational capacity for
each weight was lessened by the large architecture.

The results to this point show minimal loss in performance
(relative to Experiment \#1) after the
activation is quantized to 32 levels and 1000 weights are used.  Let us take a step-back and compare to other methods of quantization.  
Our results already improve on the absolute as well as relative loss in recall@5
accuracy seen in earlier
studies~\cite{Zhou2016,Hubara2016,wu2018training} and is close to
the best previous recall@1 loss~\cite{Zhou2016}.  However, as pointed out
in Section~\ref{weight clustering}, we might be able to do better
using a model of the expected weight distribution, rather than relying
on k-means clustering of a 2\% sample.  We chose to use a Laplacian
distribution model according to $L_1$-error spacing. Using this
approach, we see a significant improvement in performance 
(Experiment \#9).
We surpass the performance both of our fully continuous baseline
(Experiment \#1) and of our k-means weight clustering (Experiment
\#6).

Compared to the prior work that focused on moving away from
floating-point implementations (Table~\ref{table: compare}),\footnote{
  We have only included in this table those references that reported
  their work on ImageNet classification using the Alexnet
  architecture, quantizing both weights and activations, so that
  comparisons are fair.  For this reason,
  several well-known pieces of prior
  work~\cite{Vanhoucke2011,courbariaux2016,Li2017} are not included in
  Table~\ref{table: compare}.}
our approach is the only one which did not suffer a significant loss
in performance, relative to the unquantized version of the network.
We have, by far, the best performance both
relative to baseline
and in absolute terms.  \cite{nvidia2016} is the only
other reference
that we have found with weight quantization that did not suffer from
performance loss but~\cite{nvidia2016} does not quantize
activations and requires floating-point calculations.  DoReFa~\cite{Zhou2016},
which is closest to our performance,
is 8 times slower than the
baseline implementation, whereas we expect our implementation to be
as fast as or
faster than the baseline due to the relative speed of lookups versus
multiplies.

\section {Memory Savings, No Multiplication, No Floating Point}
\label{Implementation Details}

As we have shown, it is possible to train networks with little (if any)
additional error introduced through quantizing the activation
function.  On top of the quantized activation function, we can use
our clustering approach to reduce the number of unique weights in the
network.
With these two quantization components in place, an inference step
in a fully trained neural network can be completed with no
multiplication operations, no non-linear function computation, and no
floating point representations.

To accomplish this,
we have quantized the non-linear
activation function to $|A|$ activations and allowed $|W|$ unique
weights in the network.  We pre-compute all of the multiplications
required and store them in a table of size $A \times W$.  In our
AlexNet experiments, we typically used $A=32, W=1000$, which required
storing 32,000 entries.  However, this extra memory requirement is
completely offset with the savings obtained from no longer needing to
store the weights.  Previously, for each weight, a floating point
number (32 bits) was required.  With this method, only an index to the
correct \underline{column} in the table is needed (10 bits).  Given
the number of weights in a network like AlexNet ($\approx 50*10^6$),
this reflects $>69\%$ savings in memory, in addition to computational
savings detailed below.  Furthermore, in terms of bandwidth for
downloading trained models (for example to mobile devices) we can find
greater efficiency by using entropy coding of the weight indices:
based on our fully-trained quantized network weight distributions, even
the simplest (non-adaptive, marginal-only) entropy coding reduces the
index size from 10 bits to below 7 bits, giving a $> 78\%$ savings in
model storage size.\footnote{This same memory/bandwidth savings
  is available as soon as the weights are clustered, even if the activations
  are not quantized.}

\begin{figure}
  \includegraphics[width=\linewidth]{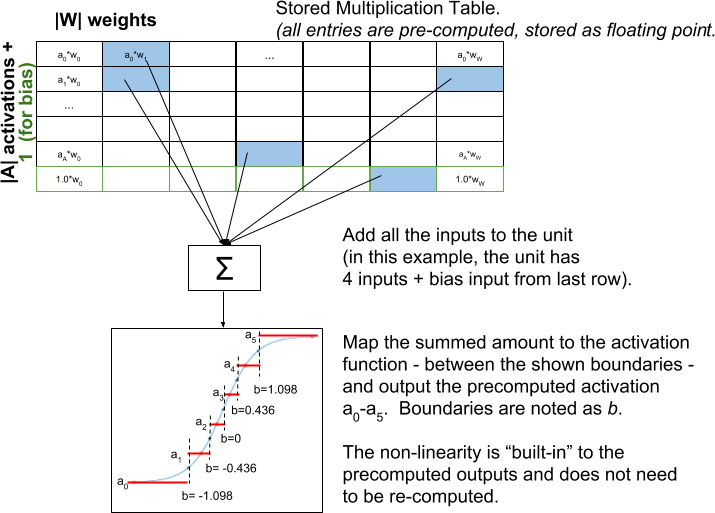}
  \caption{Using a stored multiplication table to avoid multiplies at
    inference time.  Additionally, the activation output is also stored
    and not computed; therefore no non-linearities are computed during
    inference.}
  \label {methodNoMult}

\floatspace

\end{figure}

Figure~\ref{methodNoMult} shows how a pre-computed multiplication
table could be deployed.  In the example, we
show a single unit within a network; the unit has 4 inputs + a bias
unit.  The stored multiplication table includes a row for
the bias unit's computation (e.g. multiplying the bias unit's weight
by an ``activation'' of 1.0).  Note that the same multiplication table
is used across all of the network's nodes.

Once the summation is computed by adding the looked-up entries from
the multiplication table, the activation is computed on the summed
value. In this example, the activation is shown with a tanhD
activation function with 6 levels of quantization.
We do not compute the activation function's output but instead
find the right output $j$ for $\{ a_j \}$ according to the
boundaries that we have recorded for the $b$.  This $j$ is the
\underline{row} index that is used for this value as it is fed into
the next layer's units.

There are two
remaining inefficiencies in this system.  First,
we represented all the values stored in the lookup-table (LUT) in floating point.
This has a minor impact on the total memory used but, more importantly
means we need to use a floating-point accumulator on the results.
To address this first issue, we switch to a fixed-point / integer
representation for all the stored values.
Note that all the values are
multiplied by a large scale-factor, $2^s$, to provide us with the
needed precision,
and divided by $\Delta x$, a sampling interval that we will use in the
activation input space.
The easiest method of selecting
$2^s$ is empirically, as having $s$ too large is not detrimental as
long as the additions fit in the allocated memory for the temporary
accumulator variables required.
The
summation now
emits an integer,
the activation-function input
scaled by $\frac{2^s}{\Delta x}$.

\begin{figure}
  
  \includegraphics[width=\linewidth]{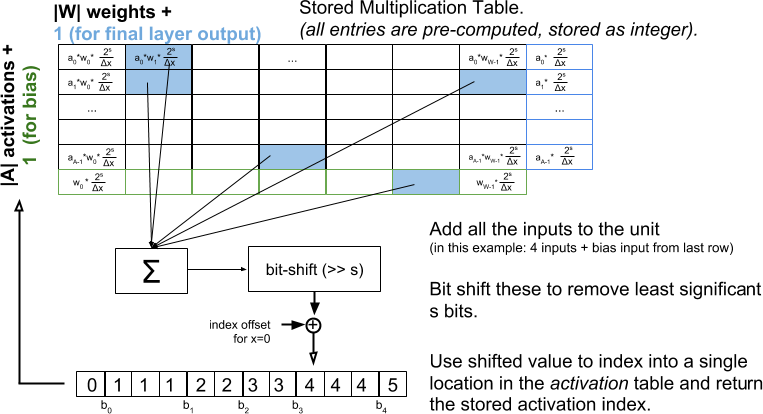}
  \caption{Extending the method shown in
    Figure~\ref{methodNoMult}. {\small
      All values in the multiplication
    table are pre-computed to include a large scale factor, $2^s$, as
    well as a activation-input quantization factor, $\frac{1}{\Delta
      x}$.
    After summing these fixed-point values (and adding the
    index offset for $x=0$), we have the index into the
    activation-function table without scanning.  To support functions
    like tanh, where the spacing between quantization boundaries is
    not uniform, we allow this quantization table to have more than
    $|A|$ entries.  Those entries simply give a value in $[0, |A|)$,
      which is the activation index for this output.  On the final
      layer, we look up the actual output value by looking into the
      column for $w = 1$. } }

  \label {methodNoMultNoFloatNoScan}    
  \floatspace

\end{figure}

The second inefficiency is related to finding the activation
function's output value (or, more accurately, the index for that value
in the LUT row space).
Using the approach from Figure~\ref{methodNoMult},
finding the 
right output (one of ${a_0-a_5}$) requires that we examine the
boundaries of the bins ($b$).
To address the inefficiency of needing to
scan
the boundaries, $b$, in the activation function, we instead directly look-up the
bin in the activation that contains the correct output (see
Figure~\ref{methodNoMultNoFloatNoScan}).
After
the summation of the inputs is computed, it is bit-shifted by $s$
bits,
removing the least significant $s$ bits and
giving the index of
the bin in the activation table to look in.
The binning (in the original activation-input space) is $\Delta
x$ wide, which we have already accounted for by the $\frac{1}{\Delta
  x}$ scale factor that we included in our stored LUT values.
So bit shifting by $s$ bits has replaced
a linear (or binary) scan of
the boundaries, giving a speed up relative to scanning (or,
alternatively, relative to general multiplication and divide
operations).

One of the limitations of this lookup-approach is that it works in
cases in which the spacings between the activation boundaries
are integer multiples of a uniform step size, $\Delta x$.
For the ReLU-6 activations, this is simple: the boundaries are spaced
uniformly; $\Delta x = \frac{6}{|A| - 1}$ and the activation table
size is just $|A|$ entries.\footnote{In fact for ReLU6 with $\Delta x
  = \frac{6}{|A| - 1}$, the activation table
  can be
  completely omitted: it is just an identity mapping.}
However, when we quantize the tanh activation
(Section~\ref{tanh}),
the width of the bins varies.  For the example shown in
Figures~\ref{methodNoMult} and~\ref{methodNoMultNoFloatNoScan}, the
quantization boundaries are adjusted slightly from the ``optimal'' (that
is, the boundary that would give the lowest quantization error
relative to the original underlying tanh nonlinearity), so that we can
set $\Delta x = 0.218$ and have an activation table that is 12 entries
long (pointing at 6 distinct quantized activation levels).  In this
case, the less we want to move our activation boundaries relative to
the expected optimal, the smaller the value that $\Delta x$ must be
and the larger that the activation table must be.  Even so, since the
table is a simple 1D array (in contrast with the multiplication
table), the amount of memory that is needed for this table is negligible.

Since we are replacing floating point computations
with table lookups and fixed-point summation, we
need to guarantee that we will not overflow our underlying integer buffers.
We are able to provide this guarantee by selecting $\frac{2^s}{\Delta
  x}$ appropriately.  Our weights, one multiplicand, are always within a
known, bounded range since we know our weight-cluster--center values.
The previous layer's outputs, the other multiplicand,
 are also
in a known, bounded range since they are one of our $|A|$
quantization levels.  Finally, by examining the network architecture, we know
the maximum number of values that will need to be added by our
summation step.
If longer integers are
required,
this increase will
only impact the buffer used by the
summation (since that is the only location impacted by the maximum
network fan-in).  In the unlikely event that longer integers are
required in the tables, this additional space
is a minor expense in
comparison to storing full resolution weights.

In summary, the final approach, shown in
Figure~\ref{methodNoMultNoFloatNoScan}, accomplishes what we set out
to do at inference time: (1) eliminate multiplications in inference
(2)
eliminate floating point in inference
and (3)
eliminate computation of non-linearities
without
scanning of an activation input-to-output array.

Note that \emph{during} training, floating point is used.
\cite{wu2018training} addresses training with integers with various
classification problems.  Our goal is to ensure that networks, even if
trained on the fastest GPUs, can be deployed on small devices that may be
constrained in terms of power, speed or memory.  Additionally, for our
requirements, which encompass deployment of networks outside of the
classification genre, we needed the quantization techniques to work
with regression/real-valued outputs.

\section {Discussion \& Future Work}

The need to enable more complex computations in the enormous number of
deployed devices, rather than sending raw signals back to remote
servers to be processed, is rapidly growing.  For example, auditory
processing within hearing aids, vision processing in remote sensors,
custom signal processing in ASICs, or any of the recent photo
applications running on the low and medium-powerful cell phones
prevalent in many developing countries, all will benefit from on-device
processing.

Pursuing quantized networks has led to a number of interesting
questions for future exploration.  Three immediate directions for
future work are presented below.  We also use this opportunity to
discuss some of the insights/trends we noticed in our study but
were not able to discuss fully here.

\vspace{-0.5\topsep}
\begin{packed_itemize}[leftmargin=.1in]
  \item For quantizing weights, all of the network's weights were
    placed into a single bucket.  An alternative is to cluster the
    weights of each layer, or set of layers, independently.  If there
    are distribution differences between layers (as can be seen in
    Figure~\ref{fig:Alexnet histograms}), this may better capture
    the most significant weights from each layer.  If this divided
    approach were used, there would be multiple multiplication tables
    stored
    for the same network.  However, for large networks, the extra
    memory requirement would still be eclipsed by the saving of not
    representing each weight individually and the order in which the
    different LUT are accessed would be predictable (based on network
    architecture), making any L1 caching more effective.

\item Currently, $|W|$ is kept constant throughout training.
  However, we have
      witnessed instability in the beginning of training
 when $|W|$ is small.
      These spikes in the training loss dissipate as
      training progresses.  Starting training with a larger-than-desired
      $|W|$ and gradually decreasing it may address
      the initial instability.

        \item We, and other studies, have noticed the
          regularization-type effects of these methods.  Additionally,
          we have noticed improved performance when other
          regularizers, such as Dropout, are not used.  The use of
          these methods as regularizers  is
          open for future work.
      
\end{packed_itemize}
\vspace{-0.5\topsep}

Beyond the practical ramifications of these simplified networks,
perhaps what is most interesting are the implications of the
simplified networks on network capacity.
In general, we have embraced training ever larger networks to address
growing task complexity and performance requirements.  However, if we
can obtain close performance using only a small fraction of the
representational power in the activations and the weights then,
with respect to our current models, much smaller
networks could store the same information.  Why does performance
improve with larger networks?  Perhaps the answer lies in the pairing
of the network architectures and the learning algorithms.  The
learning algorithms control how the search through the weight-space
progresses.  It is likely that the architectures used today are
explicitly optimized for the task and implicitly optimized for the
learning algorithms employed. The large-capacity, widely-distributed,
networks work well with the gradient descent algorithms used.
Training networks to more efficiently store information, while
maintaining performance, may require the use of alternate methods of
exploring the weight-space.

\bibliography{binary}
\bibliographystyle{sysml2019}

\end{document}